\pdfoutput=1

\documentclass[11pt]{article}

\usepackage[final]{acl}

\usepackage{times}
\usepackage{latexsym}
\usepackage{amssymb}

\usepackage[T1]{fontenc}

\usepackage[utf8]{inputenc}

\usepackage{microtype}

\usepackage{inconsolata}

\usepackage{graphicx}
\usepackage{multirow}
\usepackage{colortbl}
\usepackage{booktabs}
\usepackage{amsmath}
\definecolor{lightgray}{rgb}{0.95,0.95,0.95}

\title{IG-Pruning: Input-Guided Block Pruning for Large Language Models}

\author{
    Kangyu Qiao\textsuperscript{\rm 1,3},
    Shaolei Zhang\textsuperscript{\rm 1,3},
    Yang Feng\textsuperscript{\rm 1,2,3}\footnotemark[2] \\
    \textsuperscript{\rm 1}{Key Laboratory of Intelligent Information Processing,} \\ Institute of Computing Technology, Chinese Academy of Sciences (ICT/CAS) \\
    \textsuperscript{\rm 2} {Key Laboratory of AI Safety, Chinese Academy of Sciences} \\
    \textsuperscript{\rm 3} {University of Chinese Academy of Sciences, Beijing, China} \\
    \texttt{\{\href{mailto:qiaokangyu24s@ict.ac.cn}{qiaokangyu24s}, \href{mailto:zhangshaolei20z@ict.ac.cn}{zhangshaolei20z}, \href{mailto:fengyang@ict.ac.cn}{fengyang}\}@ict.ac.cn}
}

\begin{document}
\maketitle

\renewcommand{\thefootnote}{\fnsymbol{footnote}} 
\footnotetext[2]{Corresponding author: Yang Feng.} 
\renewcommand{\thefootnote}{\arabic{footnote}}

\begin{abstract}
With the growing computational demands of large language models (LLMs), efficient inference has become increasingly critical for practical deployment. Depth pruning has emerged as a promising approach for reducing the computational costs of large language models by removing transformer layers. However, existing methods typically rely on fixed block masks, which can lead to suboptimal performance across different tasks and inputs. In this paper, we propose IG-Pruning, a novel input-aware block-wise pruning method that dynamically selects layer masks at inference time. 
Our approach consists of two stages: (1) Discovering diverse mask candidates through semantic clustering and $L_{0}$ optimization, and (2) Implementing efficient dynamic pruning without the need for extensive training. 
Experimental results demonstrate that our method consistently outperforms state-of-the-art static depth pruning methods, making it particularly suitable for resource-constrained deployment scenarios.\footnote{\url{https://github.com/ictnlp/IG-Pruning}} \end{abstract}

\section{Introduction}

Large Language Models (LLMs) \cite{brown2020language,llama3modelcard,qwen3,zhang2024bayling,zhang2023bayling} have demonstrated remarkable capabilities across a wide range of natural language processing tasks. However, their immense model size and computational demands present significant deployment challenges \cite{wang2024model,zhou2024survey}, particularly in resource-constrained environments and for latency-sensitive real-time inference scenarios. 
To address this, pruning techniques have become a crucial area of research \cite{ma2023llm,sunSimpleEffectivePruning2023,frantar2023sparsegpt,ashkboos2024slicegpt,fangmaskllm,lingslimgpt,zhangdynamic,gu2021pruning}, being highly favored due to their potential for reducing parameters for efficient inference.

As large LLMs continue to scale in size, researchers have identified significant redundancy within their layer structures. 
Studies from  \citet{liu2023deja, men2024shortgpt, gromov2024unreasonable} reveal that word embeddings in adjacent layers often change slightly due to residual connection, suggesting that selective layer removal may have minimal impact on performance. 
These findings have motivated increasing research interest in discovering effective depth pruning strategies for LLMs, which aim to reduce the number of transformer layers or blocks in the model architecture while maintaining performance.
In recent years, depth pruning methods \cite{song2024sleb,sieberling2024evopress,kim2024shortened,lingslimgpt} have emerged as a promising approach for reducing LLM computational costs. Compared with fine-grained structured pruning methods (which remove the neurons or channels), depth pruning has demonstrated superior computational efficiency advantages in practical deployments \cite{kim2024shortened}.

\begin{figure}[t]
  \centering
  \includegraphics[width=\columnwidth]{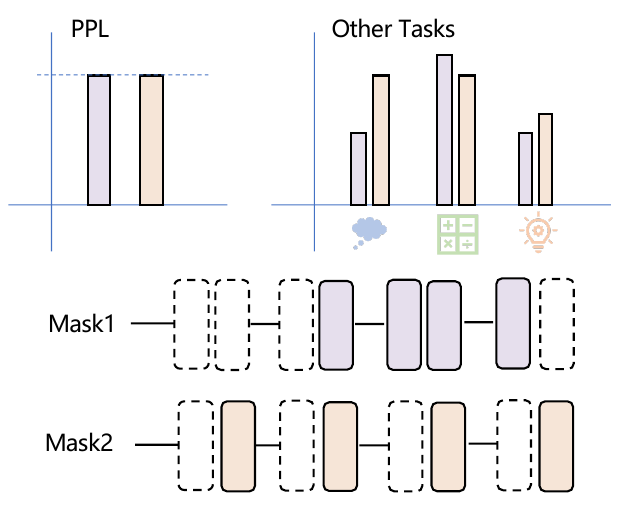}
  \caption{Different Mask structure can lead to similar perplexity scores but exhibit significant performance variations across different downstream tasks.}
  \label{fig:intro}
\end{figure}

However, a critical limitation of existing depth pruning methods is their reliance on a fixed layer pruning mask determined offline based on global layer importance metrics at a given sparsity level. This static approach is problematic because different fixed pruning masks, even at the same sparsity level, can exhibit significant performance variations across different downstream tasks. For instance, we observe that perplexity (PPL) is commonly used as a saliency metric for layer pruning \cite{sieberling2024evopress,kim2024shortened}, but as illustrated in Figure~\ref{fig:intro}, different mask structures can achieve similar perplexity scores while exhibiting substantially different performance across various downstream tasks. To overcome these limitations and enable adaptive computation pathways, researchers have explored various dynamic routing approaches \cite{elhoushi2024layerskip,fan2024not,delskipdecode,schuster2022confident,raposo2024mixture,tan2024dlo,wurouting}. However, most existing methods perform dynamic routing at the token level, which introduces significant drawbacks: they lack comprehensive understanding of sentence-level semantics, potentially leading to globally inconsistent routing decisions. Furthermore, these approaches typically incur substantial computational overhead from frequent token-level routing calls and require extensive training of additional router networks alongside the original model parameters, making them computationally expensive and time-consuming to implement.

To address the challenges identified in existing works, we propose IG-Pruning, a novel block-wise pruning method that dynamically selects layer masks based on input characteristics at inference time. Our approach consists of two stages: (1) a semantic clustering-based mask discovery stage that identifies diverse, high-quality mask candidates while capturing global information through rapidly converging trainable masks, and (2) a lightweight inference-time routing mechanism that requires no additional training of the base model parameters, enabling efficient dynamic adaptation to varying inputs.

Extensive evaluations demonstrate that our approach consistently outperforms state-of-the-art static pruning methods across different sparsity levels and model architectures on various zero-shot tasks. For Llama-3-8B at 25\% sparsity, IG-Pruning preserves 87.18\% of dense model performance, surpassing the best baseline by 10.86 percentage points. Similarly, for Qwen-3-8B, IG-Pruning maintains 96.01\% of dense model performance at 13.9\% sparsity, compared to 90.37\% for the best baseline.

Our method trains only mask parameters while keeping model weights frozen, enabling rapid adaptation with minimal computational overhead. During inference stage, it incurs negligible routing overhead by efficiently skipping unimportant layers; and these advancements provides a viable path toward deploying powerful LLMs in environments with limited computational resources.

\begin{figure*}[t]
  \centering
  \includegraphics[width=0.7\textwidth]{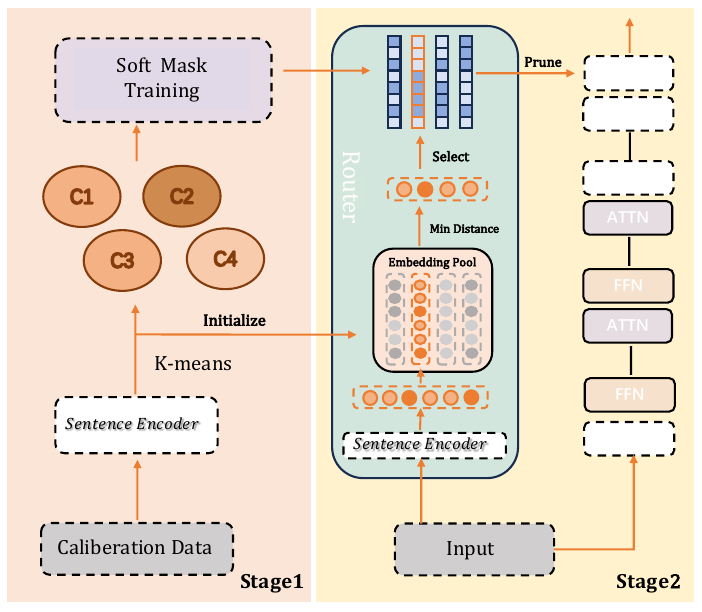}
  \caption{Overview of our method. The approach consists of two stages: (1) Preparing mask candidates through input clustering and soft mask training; (2) Dynamic pruning that selects the appropriate mask for each input at inference time. This enables efficient computation by selectively skipping layers based on input characteristics while maintaining model performance.}
  \label{fig:method_overview}
\end{figure*}

\section{Related Work}
Most static depth pruning approaches focus on calculating saliency scores for each transformer block, and removing layers according to these scores. 
Commonly used saliency metrics include cosine similarity \cite{song2024sleb, men2024shortgpt}, magnitude, second-order derivatives \cite{kim2024shortened}, and perplexity \cite{sieberling2024evopress}.
These works calculate layer importance as if they are independent of others, which ignores the coupling connections between layers.
As discovered in \citet{fan2024not}, contiguous middle layers often exhibit similar saliency scores, which inspired \citet{chen2024streamlining} to use small FFN or transformer blocks to replace contiguous layers.
EvoPress \cite{sieberling2024evopress} found that lower per-layer error does not necessarily lead to better performance, and proposed an evolutionary search algorithm to generate offspring from parent masks, then select better candidates with lower perplexity or KL divergence.
Rather than directly removing layers, LaCO \cite{yang2024laco} collapses consecutive redundant model layers via layer averaging. MKA \cite{liu2024pruning} transforms layer activations into
low-dimensional manifolds using diffusion kernel algorithms and evaluates saliency using the NPIB metric.

Beyond one-shot pruning approaches, dynamically skipping unimportant layers during inference has also emerged as a promising research direction. 
Early approaches include early skipping \cite{delskipdecode, zhu2024hierarchical}, early exit \cite{elhoushi2024layerskip}, and periodic skipping \cite{liu2024accelerating}. 
However, these methods typically require routers for each layer and demand elaborate training of original weights to recover performance. Dynamic skipping has also been adopted in long-context and multimodal models. Adaskip \cite{Adaskip_2025} focused on adaptive layer skipping for long-context models, accelerating both prefilling and decoding phases. RoE \cite{wurouting} employs token-wise routing for multimodal LLMs and trains low-rank adapters to replace the skipped layers.

\section{Method}
As illustrated in Figure~\ref{fig:method_overview}, our framework consists of two main stages: (1) Mask candidate discovery and (2) Dynamic routing. In the first stage, we cluster the semantic space of inputs and train cluster-specific masks using hard concrete distributions, resulting in diverse yet high-quality mask candidates that each specialize in handling different input patterns. During the second stage, at inference time, we employ a lightweight routing mechanism that maps each input to its most semantically similar cluster and applies the corresponding pre-trained mask, enabling efficient dynamic adaptation without requiring additional training of router networks or base model parameters.

\subsection{Stage 1: Discovering Mask Candidates}
In the first stage, we aim to discover a set of effective mask candidates for dynamic routing. Unlike existing routing methods that typically employ per-layer router networks to make skip decisions, we propose a global routing strategy that dynamically selects routing paths from a carefully curated candidate mask set. 

We design our mask candidate discovery process to satisfy two key requirements:
\textbf{Quality}: Masks must maintain strong general language generation capabilities.
\textbf{Diversity}: The candidate set must provide sufficient variety to handle different input patterns effectively.

To meet these requirements, we leverage hard concrete distribution to model transformer block masks to capture global routing information, 
and apply $L_{0}$ optimization with cluster-specific calibration data, generating masks that cover diverse computational needs.

\paragraph{Input Clustering.}
First, an encoder is used to encode each sentence $x_i$ in the calibration dataset into a fixed-dimensional embedding vector $e_i$:
\begin{equation}
e_i = \text{Encoder}(x_i)
\end{equation}
where $x_i$ represents the $i$-th input, and $e_i \in \mathbb{R}^d$, with $d$ being the dimension of the embedding vector.
Next, the K-means algorithm is applied to cluster all embedding vectors $e_1, e_2, \ldots, e_M$, where $M$ is the size of the calibration set. The K-means algorithm aims to find $N$ clusters $S = \{S_1, S_2, \ldots, S_N\}$ that minimize the within-cluster sum of squares:
\begin{equation}
\arg\min_{S} \sum_{k=1}^{N} \sum_{e_i \in S_k} \|e_i - \mu_k\|^2
\end{equation}
where $\mu_k$ is the centroid of cluster $S_k$. This results in $N$ cluster centers, each representing a class of semantically similar input sentences.

\paragraph{Mask Training.}
Hard concrete distribution \cite{louizos2018learning, xia2022structured, xia2024sheared} has been widely adopted in structured pruning. Following prior work, we incorporate hard concrete distribution to model transformer block masks, and use $L_{0}$ optimization to generate layer masks, enabling joint learning of all layer masks while incorporating global information.

For each cluster $S_k$, we train a dedicated layer mask $z^{(k)} \in \mathbb{R}^B$ using hard concrete distribution and Lagrangian sparsity, where $B$ is the total number of blocks in the model (for block-wise pruning, $B=2L$ where $L$ is the number of transformer layers, representing both attention and FFN blocks separately). Specifically, the masks $z^{(k)}$ are modeled as follows:

First, for each block $i$ in the model, sample $u^{(k)}_i$ from a uniform distribution:
\begin{equation}
u^{(k)}_i \sim \text{Uniform}(0, 1), \quad i \in \{1, 2, \ldots, B\}
\end{equation}

Then, compute the soft mask value $s^{(k)}_i$ for each block using the sigmoid function:
\begin{equation}
s^{(k)}_i = \sigma\left(\frac{1}{\beta} \log{\frac{u^{(k)}_i}{1-u^{(k)}_i}} + \log\alpha^{(k)}_i\right)
\end{equation}

Stretch the soft mask values to a specific interval $[l,r]$:
\begin{equation}
\tilde{s}^{(k)}_i = s^{(k)}_i \times (r-l) + l
\end{equation}

Finally, obtain the hardened mask $z^{(k)}_i$ for each block by clipping:
\begin{equation}
z^{(k)}_i = \min(1, \max(0, \tilde{s}^{(k)}_i))
\end{equation}

The complete mask vector for cluster $k$ is then $z^{(k)} = [z^{(k)}_1, z^{(k)}_2, \ldots, z^{(k)}_B]$, where each element corresponds to a specific transformer block in the model. During training, these mask values are soft (continuous values between 0 and 1), functioning as scaling parameters. During inference, they are binarized to either 0 (block skipped) or 1 (block executed).

Here, $\sigma$ denotes the sigmoid function. The temperature $\beta$ is fixed hyperparameter, and $l < 0, r > 0$ are two constants that stretch the sigmoid function output. $\alpha^{(k)}_i$ are the main learnable parameters for i-th block mask value in cluster $k$.

We enforce a target sparsity via a Lagrangian term. Let $s_{\text{target}}$ be the target sparsity and $t^{(k)}$ be the current sparsity of mask $z^{(k)}$ (computed as the proportion of zeroes in the mask), the Lagrangian penalty term $L_{s}^{(k)}$ is:
\begin{equation}
L_{s}^{(k)} = \lambda_1^{(k)}(t^{(k)} - s_{\text{target}}) + \lambda_2^{(k)}(t^{(k)} - s_{\text{target}})^2
\end{equation}

For the $k$-th cluster, the optimization objective for its mask parameters $\log\alpha^{(k)}$ is to minimize:
\begin{equation}
L_{\text{total}}^{(k)} = \sum_{x_j \in S_k} L_{\text{LM}}(x_j; W \odot z^{(k)}) + L_{s}^{(k)}
\end{equation}
where $L_{\text{LM}}$ is the language modeling loss and $W$ represents the model weights.

\paragraph{Routing Decision.}
To implement dynamic routing decisions, we maintain an embedding pool for each semantic cluster to represent the cluster's features. These embeddings $c_k$ are initialized using the cluster centers $\mu_k$. During inference, for each input sequence, we first extract its embedding representation $e_x$ through the encoder, then calculate the Euclidean distance between this embedding and each cluster embedding $c_k$. Based on the calculated distances, we select the most similar cluster as the best match for that input:

\begin{equation} k^* = \arg\min_k ||e_x - c_k||_2^2 , k \in \{1, 2, \ldots, N\} \end{equation}

After determining the best matching cluster, we directly adopt the trained mask corresponding to that cluster as the final execution mask for input $x$:

\begin{equation} M^{x} = z^{(k^*)} \end{equation}

where $z^{(k^*)}$ is the binary mask vector associated with cluster $k^*$, containing all block-level mask values.

\paragraph{Dynamic Routing for FFN and Attention Blocks.}
Our dynamic routing approach employs different strategies for Feed-Forward layers and Attention layers. During training, the layer mask values are soft, functioning as scaling parameters that directly multiply with the outputs of FFN and Attention components. This enables gradient-based optimization through backpropagation. 
During inference, we use hard binary masks containing only 0 and 1, where FFN layers are completely skipped when the corresponding mask value is 0. For Attention layers, the approach is more nuanced due to the necessity of maintaining key-value caches for autoregressive generation. When an Attention layer is marked for skipping, we still compute the key and value projections to maintain the KV cache, but we bypass the computationally expensive scaled dot-product operation between queries and keys. Specifically, for a transformer layer $i$ with mask value $M_{i}^{x} = 0$, the FFN computation $\text{FFN}(x_i)$ is entirely skipped, while for Attention, we compute $K = W_K x_i$ and $V = W_V x_i$ for the cache but skip $\text{Attention}(Q, K, V) = \text{softmax}(QK^T/\sqrt{d})V$. This selective computation strategy preserves the model's autoregressive capabilities while reducing computational overhead.

\section{Experiment}

\begin{table*}[t]
  \centering
  \resizebox{\textwidth}{!}{
      \begin{tabular}{l | c | l |c c c c c c | c c}
        \toprule
        \textbf{Model} & \textbf{Sparsity} & \textbf{Method} & \textbf{OBQA} & \textbf{WG} & \textbf{HS} & \textbf{PIQA} & \textbf{ARC-E} & \textbf{ARC-C} & \textbf{Average} & \textbf{Percentage} \\
        \midrule
        \multirow{13}{*}{Llama-3-8B} & 0\% & Dense & 44.6 & 73.24 & 79.16 & 80.79 & 77.82 & 53.24 & 68.14 & 100\% \\
        \cmidrule{2-11}
        & \multirow{4}{*}{12.5\%} & SLEB & 38.6 & 69.45 & 70.71 & 77.63 & 70.28 & 43.00 & 61.61 & 90.42\% \\
        & & ShortenedLlama & 39.2 & 61.56 & 66.84 & 76.33 & 67.63 & 38.57 & 58.36 & 85.64\% \\
        & & EvoPress & 41.2 & 70.17 & 72.03 & 77.75 & 71.00 & 43.69 & 62.64 & 91.93\% \\
        & &  \textbf{IG-Pruning} & \textbf{43.6} & \textbf{72.93} & \textbf{77.26} & \textbf{79.38} & \textbf{77.06} & \textbf{51.62} & \textbf{66.98} & \textbf{98.29\%} \\
        \cmidrule{2-11}
        & \multirow{4}{*}{25\%} & SLEB & 33.8 & 53.90 & 57.96 & 72.25 & 57.32 & 31.56 & 51.13 & 75.04\% \\
        & & EvoPress & 32.8 & 57.93 & 58.16 & 71.06 & 58.38 & 33.70 & 52.01 & 76.32\% \\
        & & ShortenedLlama & 33.6 & 53.91 & 57.98 & 72.31 & 57.15 & 31.74 & 51.12 & 75.01\% \\
        & &  \textbf{IG-Pruning} & \textbf{40.0} & \textbf{68.98} & \textbf{67.53} & \textbf{76.12} & \textbf{63.43} & \textbf{40.36} & \textbf{59.40} & \textbf{87.18\%} \\
        \cmidrule{2-11}
        & \multirow{4}{*}{37.5\%} & SLEB & 28.4 & 52.24 & 46.46 & 65.77 & 46.96 & 28.41 & 44.71 & 65.61\% \\
        & & EvoPress & 28.2 & 51.22 & 45.58 & 65.18 & 48.15 & 28.50 & 44.47 & 65.26\% \\
        & & ShortenedLlama & 28.6 & 52.41 & 45.90 & 64.69 & 42.68 & 27.47 & 43.63 & 64.02\% \\
        & &  \textbf{IG-Pruning} & \textbf{31.8} & \textbf{58.01} & \textbf{49.63} & \textbf{65.94} & \textbf{48.44} & \textbf{30.38} & \textbf{47.37} & \textbf{69.51\%} \\
        \midrule
        \multirow{13}{*}{Qwen-3-8B} & 0\% & Dense & 41.8 & 67.96 & 74.93 & 77.48 & 80.77 & 56.40 & 66.56 & 100\% \\
        \cmidrule{2-11}
        & \multirow{4}{*}{13.9\%} & SLEB & 37.4 & 60.85 & 62.45 & 77.52 & 74.45 & 47.09 & 59.96 & 90.09\% \\
        & & ShortenedLlama & 37.0 & 59.27 & 61.82 & 75.14 & 71.00 & 45.14 & 58.23 & 87.49\% \\
        & & EvoPress & 39.0 & 61.96 & 67.76 & 75.57 & 70.33 & 46.25 & 60.15 & 90.37\% \\
        & &  \textbf{IG-Pruning} & \textbf{39.8} & \textbf{65.82} & \textbf{69.44} & 77.09 & \textbf{77.35} & \textbf{53.92} & \textbf{63.90} & \textbf{96.01\%} \\
        \cmidrule{2-11}
        & \multirow{4}{*}{25\%} & SLEB & 36.6 & 56.35 & 53.95 & 72.47 & 65.36 & 37.20 & 53.66 & 80.62\% \\
        & & EvoPress & 37.0 & 58.08 & 57.18 & 71.43 & 62.28 & 38.65 & 54.10 & 81.29\% \\
        & & ShortenedLlama & 35.6 & 53.99 & 52.20 & 70.84 & 64.69 & 36.43 & 52.29 & 78.56\% \\
        & &  \textbf{IG-Pruning} & 35.6 & \textbf{60.46} & \textbf{61.65} & \textbf{73.39} & \textbf{68.94} & \textbf{44.80} & \textbf{57.47} & \textbf{86.35\%} \\
        \cmidrule{2-11}
        & \multirow{4}{*}{36.1\%} & SLEB & 29.6 & 52.40 & 44.02 & 65.77 & 51.68 & 31.39 & 45.81 & 68.82\% \\
        & & EvoPress & 31.6 & 52.17 & 45.29 & 62.95 & 51.09 & 29.18 & 45.38 & 68.18\% \\
        & & ShortenedLlama & 28.2 & 50.91 & 37.08 & 61.75 & 46.13 & 25.43 & 41.58 & 62.48\% \\
        & & \textbf{IG-Pruning} & \textbf{32.6} & \textbf{53.43} & \textbf{49.17} & \textbf{65.83} & \textbf{54.21} & \textbf{32.17} & \textbf{47.90} & \textbf{71.96\%} \\
        \bottomrule 
      \end{tabular}
  }
  \caption{Zero-shot evaluation results on Llama-3-8B and Qwen-3-8B across multiple sparsity levels.}
  \label{tab:main_results}
\end{table*}

\subsection{Experimental Setup}
\paragraph{Datasets and Evaluation Metrics.}
Following prior work, we use lm-evaluation-harness \cite{gao2023framework} to evaluate our method on six widely-used zero-shot tasks: OpenBookQA \cite{mihaylov2018obqa}, which tests elementary-level science reasoning requiring the combination of facts with commonsense knowledge; 
Winogrande \cite{sakaguchi2021winogrande}, a large-scale adversarial dataset for testing pronoun disambiguation through commonsense reasoning; 
HellaSwag \cite{zellers2019hellaswag}, which challenges models to select plausible scenario completions through commonsense inference; 
PIQA \cite{bisk2020piqa}, focused on physical commonsense knowledge; and the ARC dataset \cite{clark2018think}, divided into ARC-Easy and ARC-Challenge subsets for testing scientific reasoning at different difficulty levels.
Llama-3-8B \cite{llama3modelcard} and Qwen-3-8B \cite{qwen3} are used as our base models, and we use all-MiniLM-L6-v2 from sentence transformer \cite{reimers2019sentence} as sentence encoder.
For calibration data for clustering and layer mask training, we use \texttt{fineweb-edu} \cite{lozhkov2024fineweb-edu}, which contains high quality synthetic data used for LLM pretraining.
\paragraph{Baselines and Setups.}
To evaluate our dynamic block pruning approach against static methods, we select three representative block pruning techniques for comparison:
\begin{itemize}
    \item \textbf{SLEB} \cite{song2024sleb}: A method that iteratively eliminates redundant transformer blocks based on cosine similarity between adjacent layers.
    
    \item \textbf{ShortenedLlama} \cite{kim2024shortened}: An approach that uses magnitude, second-order derivatives, or perplexity to measure block-level importance. After identifying unimportant blocks, this method removes them in a single pass.
    
    \item \textbf{EvoPress} \cite{sieberling2024evopress}: A technique leveraging evolutionary algorithms to search for optimal pruning masks with improved perplexity or KL divergence. 
    Starting with a random initial configuration, in each generation it mutates the compression levels of selected layers and retains the best candidates according to a fitness function. This approach yields better results but incurs higher computational costs.
\end{itemize}

For all baseline methods, we perform one-shot pruning that identifies and eliminates redundant transformer blocks without retraining, and we use \texttt{wikitext2} \cite{merity2016pointer} as calibration set for baselines.

\subsection{Main Results}

IG-Pruning consistently outperforms all baseline methods across all evaluated sparsity configurations for both Llama-3-8B and Qwen-3-8B models. 
In this paper, the sparsity level is defined as the ratio of the number of skipped blocks to the total number of blocks in the model.
For Llama-3-8B at 12.5\% sparsity, IG-Pruning maintains 98.29\% of the dense model performance, surpassing the best baseline (EvoPress) by 6.36 percentage points. This advantage becomes even more significant at 25\% sparsity, where IG-Pruning achieves 87.18\% of dense performance compared to the best baseline at 76.32\%, representing a 10.86 percentage point improvement. 
Similarly, for Qwen-3-8B, IG-Pruning preserves 96.01\% of dense model performance at 13.9\% sparsity, compared to 90.37\% for the best baseline. These consistent improvements across different model architectures demonstrate the inherent advantage of our dynamic routing strategy over static pruning methods.

\subsection{Analysis}
\paragraph{Mask Training Efficiency.}
In Stage 1 of our approach, model parameters remain frozen while only layer mask parameters undergo optimization. We set a higher learning rate for $L_{0}$ module, enabling rapid mask convergence without extensive training periods.
For our experiments, we sample 1,000 examples from each cluster for training, utilizing 4 NVIDIA H800 GPUs. Hyperparameters can be found in Appendix~\ref{tab:hyperparameters}.
For configurations with sparsity levels below 25\% across 16 clusters, all masks can be trained in approximately 15 minutes. Higher sparsity (37\%) requires around one hour of training time for mask convergence.
Our method requires training, but it only trains the block mask parameters, while the parameters in the original models are frozen. Therefore, it doesn't require excessive memory, which has been tested successfully on a single RTX 3090 for 8B model.

\begin{figure}[t]
  \centering
  \includegraphics[width=\columnwidth]{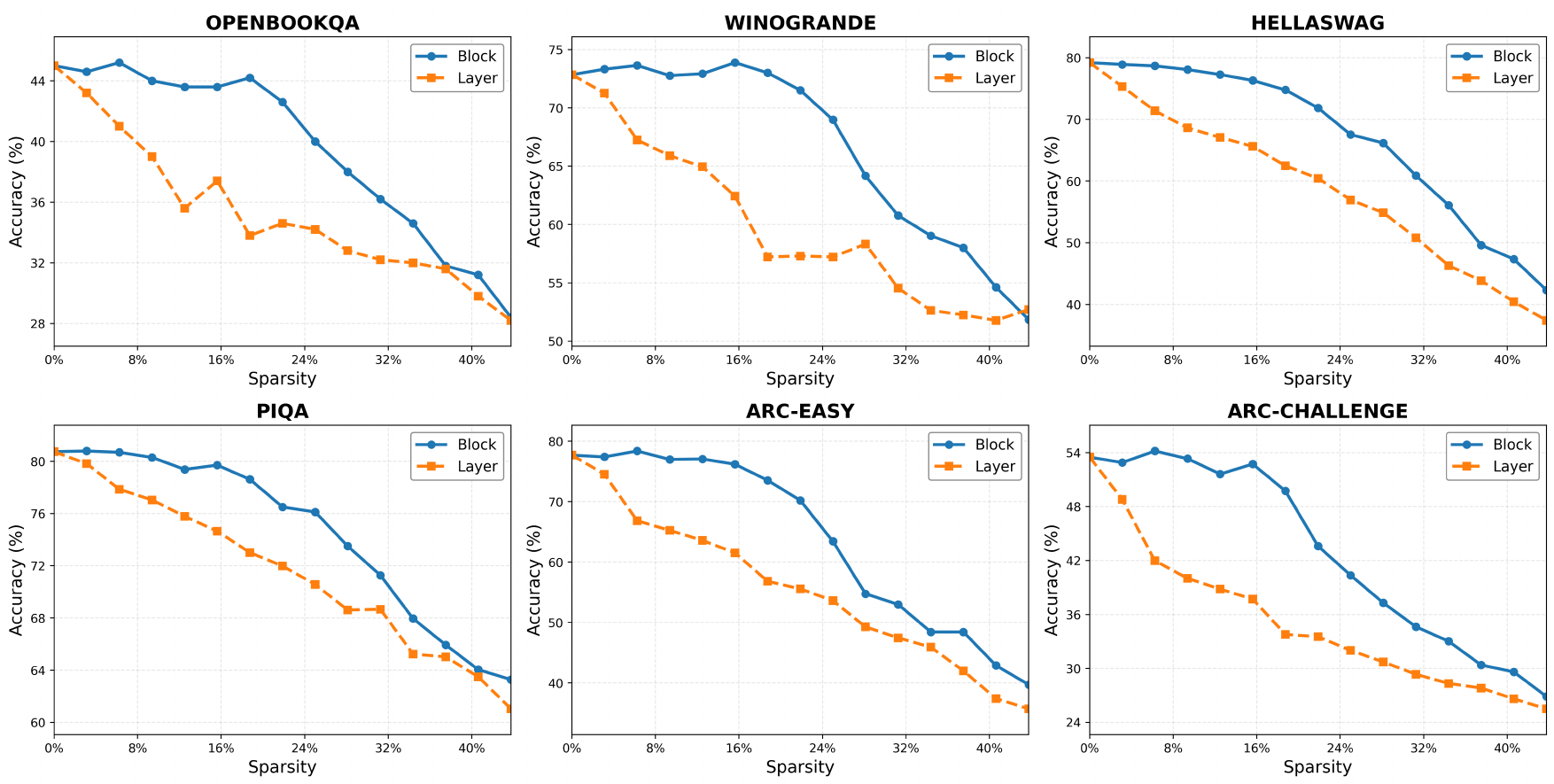}
  \caption{Results on average zero-shot task performance of Llama-3-8B, with block and layer pruning.}
  \label{fig:ablation_granularity}
\end{figure}

\paragraph{Block-level vs. Layer-level Pruning.}
To investigate the impact of pruning granularity on model performance, we conducted comprehensive experiments comparing block-level and layer-level pruning across different sparsity configurations. 
As shown in Figure~\ref{fig:ablation_granularity}, block-level pruning consistently outperforms layer-level pruning across all tasks, with performance advantages that vary based on sparsity levels. 
The gap between these approaches is most significant at sparsity levels around 20\%, where block pruning demonstrates substantially better performance. 
This suggests that independently pruning Attention and FFN components provides the model with greater flexibility to maintain critical capabilities while reducing computational costs.

\begin{figure*}[t]
  \centering
  \includegraphics[width=\textwidth]{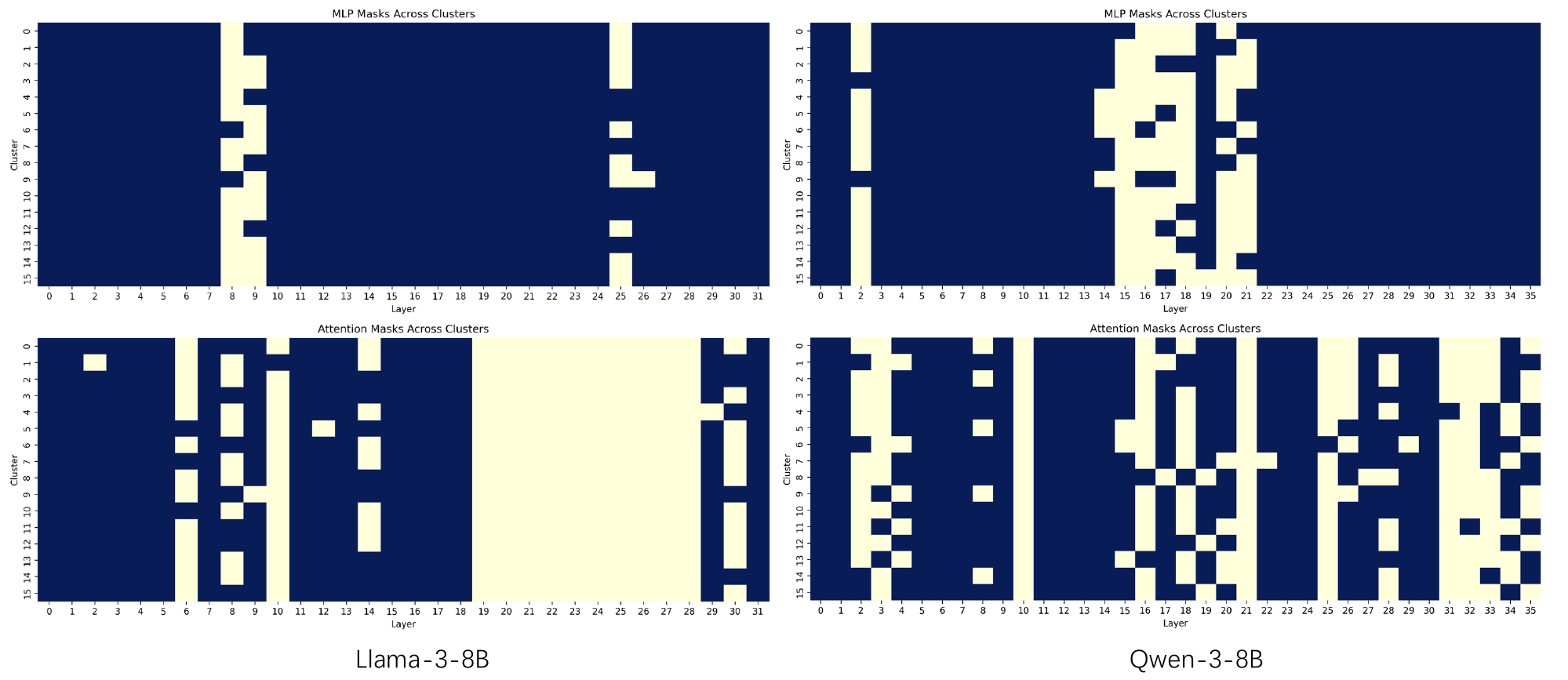}
  \caption{Block mask visualization of Llama-3-8B(left) and Qwen-3-8B(right) with 16 clusters and 25\% sparsity. Upper part is FFN Block and the lower part is Attention Block. The color indicates the mask value, with 1 being blue and 0 being yellow. }
  \label{fig:ablation_mask}
\end{figure*}

Interestingly, the performance differential diminishes as sparsity decreases. At sparsity levels higher than 40\%, the differences become minimal, and in specific tasks such as Winogrande, layer-level pruning occasionally outperforms block-level pruning. 
To better understand the results, we analyze the layer masks. Visualization in Figure~\ref{fig:ablation_mask} reveals that Llama attention blocks are more likely to be pruned compared to FFN blocks, especially in middle layers, aligning with previous observations about layer representation similarity in \citet{men2024shortgpt}.
This phenomenon also exists in the Qwen-3 model, but shows a more balanced distribution between attention and FFN blocks. Additionally, attention masks are more separate for Qwen, with no long ranges of consecutive blocks being masked. We analyzed the mask distribution at various sparsity levels and found this phenomenon was commonly observed.
This suggests that, in higher sparsity settings, retaining the FFN blocks is more beneficial for model performance, as they are more likely to contain important information. For higher sparsity levels, more FFN blocks are pruned, leading to similar performance between block-level and layer-level pruning.

\paragraph{Computational Efficiency Analysis.}
To quantify efficiency improvements, we measured FLOPs (floating point operations) for Llama-3-8B with different sparsity settings, as shown in Table~\ref{tab:block_flops}.
Our analysis reveals that block-wise pruning provides significant computational savings while maintaining model performance. At 25\% sparsity, our approach reduces the computational cost to 89.8\% of the dense model, representing a reasonable trade-off between efficiency and effectiveness. As sparsity increases to 37.5\%, computational requirements drop to 75.8\% of the original model.

\begin{table}[h]
  \centering
  \small
  \setlength{\tabcolsep}{4pt}
  \begin{tabular}{c|c|c||c|c|c}
    \hline
    \textbf{Sparsity} & \textbf{FLOPs} & \textbf{Per\%} & \textbf{Sparsity} & \textbf{FLOPs} & \textbf{Per\%} \\
    \hline
    0\% & 32.94T & 100.0\% & 21.88\% & 31.01T & 94.1\% \\
    3.12\% & 32.66T & 99.1\% & 25.00\% & 29.57T & 89.8\% \\
    6.25\% & 32.39T & 98.3\% & 28.12\% & 28.71T & 87.2\% \\
    9.38\% & 32.11T & 97.5\% & 31.25\% & 27.27T & 82.8\% \\
    12.50\% & 31.84T & 96.7\% & 34.38\% & 26.41T & 80.2\% \\
    15.62\% & 31.56T & 95.8\% & 37.50\% & 24.97T & 75.8\% \\
    18.75\% & 31.29T & 95.0\% & 40.62\% & 24.69T & 74.9\% \\
    \hline
  \end{tabular}
  \caption{Computational efficiency at different sparsity for block-wise pruning. The FLOPs values represent the computational cost, while the percentage shows the proportion relative to the dense model.}
  \label{tab:block_flops}
\end{table}


\subsubsection{Analyze clustering effectiveness}
\begin{figure}
  \centering
  \includegraphics[width=\columnwidth]{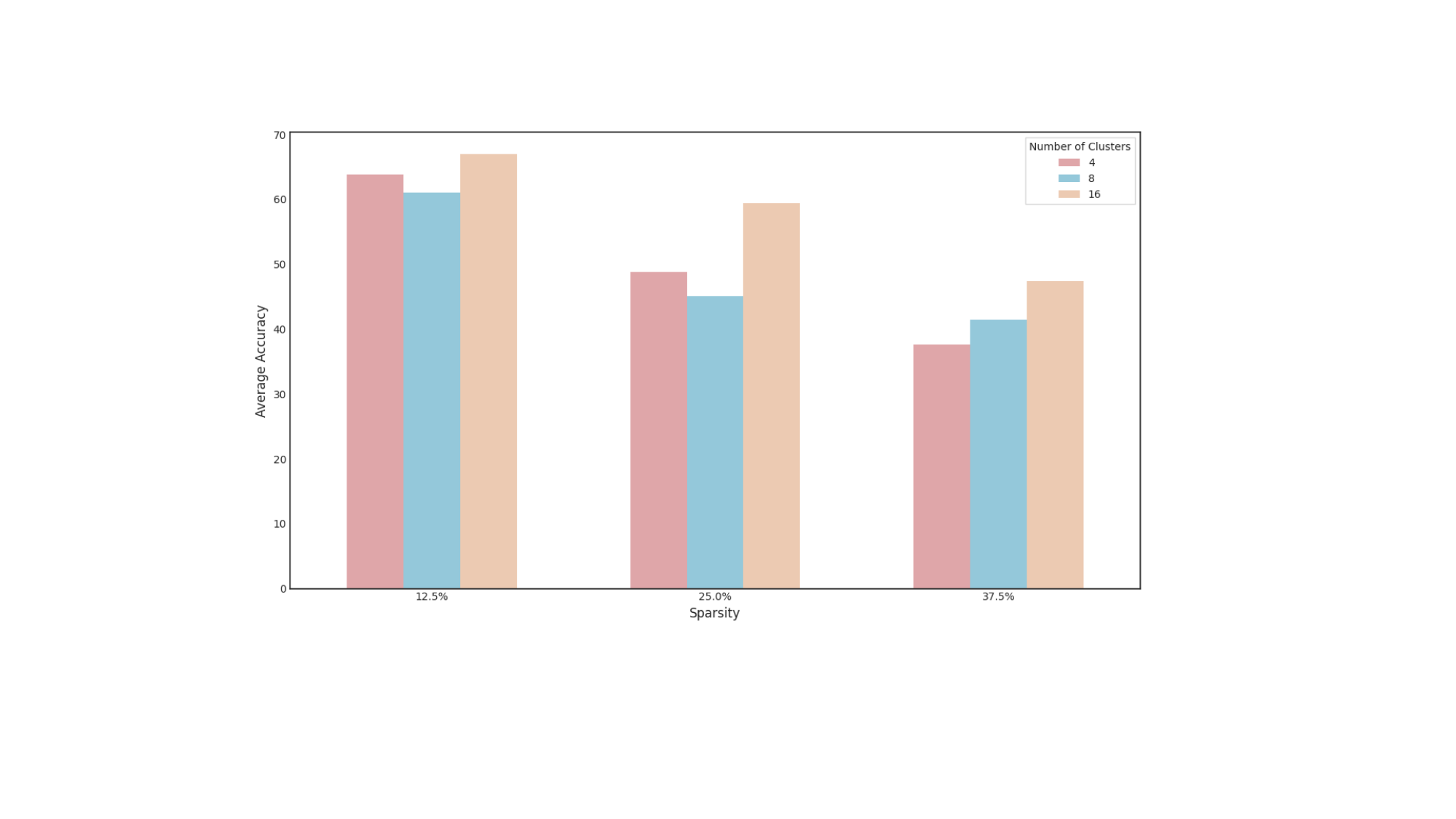}
  \caption{Impact of cluster number on performance across evaluation tasks. Results on average zero-shot task performance on Llama-3-8B, with cluster N=4, 8, and 16.}
  \label{fig:ablation_clusters}
\end{figure}

\paragraph{Number of Clusters.}
To investigate how the number of clusters affects model performance, we conducted experiments with varying cluster counts (N = 4, 8, 16) at different sparsity levels, as shown in Figure~\ref{fig:ablation_clusters}. The results demonstrate a clear trend: as the number of clusters increases, overall performance improves consistently across all pruning configurations. 
At lower sparsity, models with 16 clusters achieve an average performance of 66.98\%, compared to 61.05\% with 8 clusters and 63.82\% with 4 clusters.
This advantage becomes even more pronounced at higher sparsity levels. With sparsity of 37.5\%, the 16-cluster configuration outperforms the 4-cluster variant by 10.64 percentage points.
This pattern confirms that a higher number of clusters enables more specialized mask combinations tailored to different input types. With more clusters, the model can develop a more diverse set of computational paths, each optimized for specific semantic patterns in the input data. The performance improvements with increased cluster count provide strong evidence supporting our hypothesis that dynamic routing significantly benefits model effectiveness by enabling adaptive computation. 
Rather than forcing all inputs through a single pruned structure, our approach leverages the complementary strengths of mask combinations, explained why our dynamic pruning strategy consistently outperforms static pruning methods.

\paragraph{Calibration Data Quality.}
The quality of calibration data proves critical for effective mask training, as demonstrated in our ablation studies (Table~\ref{tab:calib_ablation}). We found that using high-quality, diverse pretraining data from \texttt{fineweb-edu} \cite{lozhkov2024fineweb-edu} yields the best results, achieving an average score of 59.40. 
In contrast, using \texttt{wikitext2}, the calibration dataset for baseline models, leads to a significant performance degradation, with the average score dropping to 55.85. Also, instruction dataset in \citet{gou2023mixture}, achieved a competitive score of 58.20 but was still lower than \texttt{fineweb-edu}. Our experiments demonstrate that clustering semantically-rich texts creates more meaningfully differentiated clusters, enabling the discovery of truly specialized computational paths. This finding highlights the importance of data diversity and representational richness in training effective dynamic routing mechanisms.

\begin{table}
    \centering
    \resizebox{\columnwidth}{!}{
        \begin{tabular}{llccccccc}
        \toprule
        \textbf{Dataset} & \textbf{OBQA} & \textbf{WG} & \textbf{HS} & \textbf{PIQA} & \textbf{ARC-E} & \textbf{ARC-C} & \textbf{Average}\\
        \midrule
        Instruction & 36.4 & 68.27 & 68.14 & 73.06 & 63.38 & 39.93 & 58.20 \\
        Wikitext2 & 39.0 & 63.06 & 64.12 & 73.07 & 60.19 & 35.67 & 55.85 \\
        Fineweb-edu & 40.0 & 68.98 & 67.53 & 76.12 & 63.43 & 40.36 & 59.40 \\
        \bottomrule
        \end{tabular}
    }
    \caption{Ablation results on Llama-3-8B with 25\% sparsity across different datasets. Comparing with \texttt{fineweb-edu}, instruction set show minor difference, while wikitext cause average score degradation.}
    \label{tab:calib_ablation}
\end{table}

To verify that the observed performance enhancement is attributable to our proposed method rather than the calibration data, we benchmarked the SLEB baseline on both the \texttt{wikitext2} and \texttt{fineweb-edu} datasets. As detailed in Table~\ref{tab:calib_baseline}, the baseline's performance did not improve when using \texttt{fineweb-edu}. Crucially, our method continues to outperform the baseline even when using \texttt{wikitext2}. This evidence indicates that the performance gains originate from our method's dynamic architecture and its ability to leverage high-quality data, rather than from an unfair data advantage.

\begin{table}
\centering
    \resizebox{\columnwidth}{!}{
        \begin{tabular}{llccccccc}
        \toprule
        \textbf{Method} & \textbf{Dataset} & \textbf{OBQA} & \textbf{WG} & \textbf{HS} & \textbf{PIQA} & \textbf{ARC-E} & \textbf{ARC-C} & \textbf{Average}\\
        \midrule
        SLEB & Wikitext2 & 33.8 & 53.95 & 57.96 & 72.25 & 57.32 & 31.56 & 51.13 \\
        SLEB & Fineweb-edu & 33.0 & 52.56 & 57.19 & 72.79 & 56.60 & 32.84 & 50.83 \\
        \midrule
        IG-Pruning & Wikitext2 & 39.0 & 63.06 & 64.12 & 73.07 & 60.19 & 35.67 & 55.85 \\
        IG-Pruning & Fineweb-edu & 40.0 & 68.98 & 67.53 & 76.12 & 63.43 & 40.36 & 59.40 \\
        \bottomrule
        \end{tabular}
    }
    \caption{Comparison with baseline models on different datasets. Our method outperforms the baseline (SLEB) regardless of the dataset used.}
    \label{tab:calib_baseline}
\end{table}


\section{Conclusion}
We introduced IG-Pruning, a novel approach for efficient LLM inference through input-adaptive dynamic block pruning. Our method addresses critical limitations of static pruning, and demonstrates that IG-Pruning consistently outperforms state-of-the-art static pruning methods across various configurations and model architectures.
Our approach offers four key advantages: (1) improved accuracy through input-adaptive computation that tailors pruning decisions to specific input characteristics, (2) efficient training that keeps model weights frozen while only optimizing lightweight mask parameters, (3) minimal inference overhead via a simple yet effective semantic-based routing mechanism, and (4) flexible block-level pruning granularity that allows independent treatment of attention and FFN components.
The success of IG-Pruning highlights the importance of input-adaptive computation in efficient LLM deployment and represents a promising direction for developing high-performing LLMs for resource-constrained environments.

\section*{Limitations}
The performance heavily depends on clustering quality, potentially diminishing if semantic clusters aren't effectively differentiated. 
Moreover, the result is sensitive to calibration data quality, as instruction datasets led to performance degradation compared to diverse pretraining data. 
Also, our evaluation focused primarily on specific zero-shot tasks, leaving generalization to other task types or domain-specific applications less thoroughly validated. 
Additionally, the method introduces sensitivity to multiple hyperparameters, including $L_{0}$ regularization, lagrangian parameters, and cluster numbers.
Finally, our work does not investigate the impact of block pruning on model factuality. Removing computational blocks risks eliminating components that are critical for factual recall, which may increase the model's propensity for hallucination. A promising direction for future work would be to combine our dynamic pruning strategy with hallucination mitigation techniques. For instance, integrating methods like TruthX \cite{zhang2024truthx}, which enhances truthfulness by editing internal model representations, or Truth-Aware Context Selection \cite{yu2024truth}, which filters untruthful information from the input context. Such an approach could lead to models that are not only efficient but also more robust and factually reliable. 

\section*{Acknowledgements}
We thank all the anonymous reviewers for their insightful and valuable comments on this paper. This work was supported by the grant from the National Natural Science Foundation of China (No. 62376260).

\bibliography{custom}

\appendix
\label{sec:appendix}
\section{Hyperparameter}
The hyperparamters we use in our experiments are listed in Table~\ref{tab:hyperparameters}.

\begin{table}[h]
  \centering
  \small
  \begin{tabular}{l|c}
    \hline
    \textbf{Hyperparameter} & \textbf{Value} \\
    \hline
    Batch Size & 32 \\
    $L_{0}$ Module Learning Rate & 0.1 \\
    Lagrangian Learning Rate & 0.1 \\
    $\epsilon$ & 1e-6 \\
    $1/\beta$ & 2/3 \\
    $l$ & -0.1 \\
    $r$ & 1.1 \\
    Number of Clusters & 16, 8, 4 \\
    Calibration Data Size for each cluster & 1000 \\
    Clustering Stage Sequence Length & 4096 \\
    Mask Training Sequence Length & 512 \\
    \hline
  \end{tabular}
  \caption{Hyperparameters used in our experiments.}
  \label{tab:hyperparameters}
\end{table}

\section{More results on various models}

To further validate the generalizability and robustness of our approach, we conducted additional experiments on a wider range of models, including Llama-3.2-3B (Table~\ref{tab:llama32-3b}), Llama-3.2-1B (Table~\ref{tab:llama32-1b}), and Qwen-3-4B (Table~\ref{tab:qwen3-4b}).
Across all tested models and architectures, the input-adaptive nature of IG-Pruning allows it to retain significantly more of the original model's performance compared to baselines, especially at moderate sparsity levels. As sparsity becomes extremely high, the performance of both methods naturally converges. These comprehensive results validate that our dynamic approach is a consistently superior and more robust solution for model pruning.

\begin{table*}
\centering
\resizebox{\textwidth}{!}{
\begin{tabular}{@{}llccccccccc@{}}
\toprule
\textbf{Model} & \textbf{Sparsity} & \textbf{Method} & \textbf{OpenBookQA} & \textbf{Winogrande} & \textbf{Hellaswag} & \textbf{PIQA} & \textbf{ARC-E} & \textbf{ARC-C} & \textbf{Average} & \textbf{Percentage(\%)} \\ \midrule
\multirow{7}{*}{Llama-3.2-3B} & 0\% (0/28) & Dense & 43.20 & 69.38 & 73.73 & 77.27 & 71.84 & 45.99 & 63.55 & 100\% \\ \cmidrule(l){2-11} 
 & \multirow{2}{*}{14\% (4/28)} & SLEB & 35.80 & 58.45 & 58.20 & 73.12 & 57.02 & 33.70 & 52.71 & 82.94\% \\
 & & IG-Pruning & 41.40 & 66.45 & 68.20 & 75.95 & 68.13 & 43.34 & 60.58 & 95.32\% \\ \cmidrule(l){2-11} 
 & \multirow{2}{*}{25\% (7/28)} & SLEB & 25.00 & 53.82 & 46.67 & 68.28 & 50.96 & 29.01 & 46.79 & 73.63\% \\
 & & IG-Pruning & 36.40 & 57.76 & 60.14 & 71.87 & 54.88 & 33.19 & 52.36 & 82.40\% \\ \cmidrule(l){2-11} 
 & \multirow{2}{*}{39\% (11/28)} & SLEB & 26.80 & 51.06 & 37.26 & 61.58 & 40.02 & 24.65 & 40.23 & 63.30\% \\
 & & IG-Pruning & 28.00 & 49.83 & 38.52 & 61.53 & 38.17 & 24.40 & 40.07 & 63.05\% \\ \bottomrule
\end{tabular}
}
\caption{Results on Llama-3.2-3B.}
\label{tab:llama32-3b}
\end{table*}

\begin{table*}
\centering
\resizebox{\textwidth}{!}{
\begin{tabular}{@{}llccccccccc@{}}
\toprule
\textbf{Model} & \textbf{Sparsity} & \textbf{Method} & \textbf{OpenBookQA} & \textbf{Winogrande} & \textbf{Hellaswag} & \textbf{PIQA} & \textbf{ARC-E} & \textbf{ARC-C} & \textbf{Average} & \textbf{Percentage(\%)} \\ \midrule
\multirow{7}{*}{Llama-3.2-1B} & 0\% (0/16) & Dense & 37.40 & 60.36 & 63.64 & 74.43 & 60.27 & 36.26 & 55.38 & 100\% \\ \cmidrule(l){2-11} 
 & \multirow{2}{*}{12.5\% (2/16)} & SLEB & 30.60 & 55.16 & 48.74 & 68.55 & 48.48 & 28.41 & 46.66 & 84.24\% \\
 & & IG-Pruning & 35.00 & 60.45 & 59.65 & 72.79 & 57.32 & 33.87 & 53.18 & 96.02\% \\ \cmidrule(l){2-11} 
 & \multirow{2}{*}{25\% (4/16)} & SLEB & 27.80 & 51.63 & 37.50 & 63.11 & 40.19 & 23.72 & 40.65 & 73.40\% \\
 & & IG-Pruning & 27.00 & 54.78 & 40.30 & 62.08 & 40.24 & 27.22 & 41.94 & 75.72\% \\ \cmidrule(l){2-11} 
 & \multirow{2}{*}{37.5\% (6/16)} & SLEB & 27.00 & 49.88 & 29.90 & 56.03 & 30.93 & 22.01 & 35.96 & 64.93\% \\
 & & IG-Pruning & 24.40 & 50.98 & 30.90 & 56.31 & 30.72 & 25.08 & 36.40 & 65.72\% \\ \bottomrule
\end{tabular}
}
\caption{Results on Llama-3.2-1B.}
\label{tab:llama32-1b}
\end{table*}

\begin{table*}
\centering
\resizebox{\textwidth}{!}{
\begin{tabular}{@{}llccccccccc@{}}
\toprule
\textbf{Model} & \textbf{Sparsity} & \textbf{Method} & \textbf{OpenBookQA} & \textbf{Winogrande} & \textbf{Hellaswag} & \textbf{PIQA} & \textbf{ARC-E} & \textbf{ARC-C} & \textbf{Average} & \textbf{Percentage(\%)} \\ \midrule
\multirow{7}{*}{Qwen-3-4B} & 0\% (0/36) & Dense & 40.40 & 65.82 & 68.42 & 75.13 & 53.75 & 53.75 & 59.55 & 100\% \\ \cmidrule(l){2-11} 
 & \multirow{2}{*}{14\% (5/36)} & SLEB & 35.40 & 56.19 & 57.36 & 72.85 & 65.78 & 39.84 & 54.57 & 91.64\% \\
 & & IG-Pruning & 37.60 & 62.58 & 59.76 & 73.55 & 68.35 & 44.62 & 57.74 & 96.97\% \\ \cmidrule(l){2-11} 
 & \multirow{2}{*}{25\% (9/36)} & SLEB & 32.20 & 53.03 & 46.94 & 67.46 & 58.37 & 31.22 & 48.20 & 80.95\% \\
 & & IG-Pruning & 35.80 & 56.43 & 53.78 & 69.85 & 60.01 & 39.07 & 52.49 & 88.15\% \\ \cmidrule(l){2-11} 
 & \multirow{2}{*}{36\% (13/36)} & SLEB & 29.80 & 53.43 & 39.54 & 62.67 & 47.01 & 26.79 & 43.21 & 72.56\% \\
 & & IG-Pruning & 30.60 & 54.69 & 42.74 & 63.65 & 47.26 & 28.66 & 44.60 & 74.90\% \\ \bottomrule
\end{tabular}
}
\caption{Results on Qwen-3-4B.}
\label{tab:qwen3-4b}
\end{table*}

\end{document}